\documentclass[sigconf]{acmart}

\AtBeginDocument{%
  \providecommand\BibTeX{{%
    \normalfont B\kern-0.5em{\scshape i\kern-0.25em b}\kern-0.8em\TeX}}}

\setcopyright{acmcopyright}
\copyrightyear{2024}
\acmYear{2024}
\acmDOI{XXXXXXX.XXXXXXX}

\acmConference[SIGCSE '24]{}{March 20--23,
  2024}{Portland, OR}
\acmPrice{15.00}
\acmISBN{978-1-4503-XXXX-X/18/06}

\usepackage{enumitem}
\usepackage{times}
\usepackage{latexsym}
\usepackage{fontawesome}
\usepackage{pifont}
\usepackage{listings}[language=Python]
\usepackage{graphicx}
\usepackage{booktabs}
\usepackage{mdframed}
\usepackage{lipsum}
\usepackage{placeins}
\usepackage{multirow}
\usepackage{amsmath}
\usepackage{hyperref}
\usepackage{cancel}
\usepackage{array}
\usepackage{xspace}

\definecolor{mygray}{gray}{0.97}
\colorlet{shadecolor}{mygray}

\newmdenv[%
  backgroundcolor=mygray, 
  linewidth=0pt
]{newshaded}

\definecolor{codegreen}{rgb}{0,0.6,0}
\definecolor{codegray}{rgb}{0.5,0.5,0.5}
\definecolor{codepurple}{rgb}{0.58,0,0.82}
\definecolor{backcolour}{rgb}{0.95,0.95,0.92}

\definecolor{ForestGreen}{HTML}{029147}

\lstdefinestyle{mystyle}{
    language=Python,
    backgroundcolor=\color{backcolour},   
    commentstyle=\color{codegreen},
    keywordstyle=\color{blue},
    numberstyle=\tiny\color{codegray},
    stringstyle=\color{codepurple},
    basicstyle=\ttfamily\footnotesize,
    breakatwhitespace=false,         
    breaklines=true,                 
    captionpos=b,                    
    keepspaces=true,                 
    numbers=left,                    
    numbersep=5pt,                  
    showspaces=false,                
    showstringspaces=false,
    showtabs=false,                  
    tabsize=2,
    columns=fullflexible,
}

\newcommand{\tcr}[1]{\textcolor{red}{#1}}
\newcommand{\tcb}[1]{\textcolor{blue}{#1}}
\newcommand{\tcg}[1]{\textcolor{ForestGreen}{#1}}

\begin{document}


\title[Can Language Models Employ the Socratic Method?]{Can Language Models Employ the Socratic Method? Experiments with Code Debugging}


\author{Erfan Al-Hossami, Razvan Bunescu, Justin Smith}
\email{ealhossa,rbunescu,jsmit840@uncc.edu}
 \orcid{0000-0002-8436-8974}
\affiliation{%
  \institution{UNC Charlotte}
  \streetaddress{9201 University City Blvd}
  \city{Charlotte}
  \state{North Carolina}
  \country{USA}
  \postcode{28213}
}


\author{Ryan Teehan}
\email{rst306@nyu.edu}
\affiliation{%
  \institution{New York University}
  \city{New York}
  \state{New York}
  \country{USA}
}

\renewcommand{\shortauthors}{Al-Hossami, et al.}

\begin{abstract}
When employing the Socratic method of teaching, instructors guide students toward solving a problem on their own rather than providing the solution directly. While this strategy can substantially improve learning outcomes, it is usually time-consuming and cognitively demanding. Automated Socratic conversational agents can augment human instruction and provide the necessary scale, however their development is hampered by the lack of suitable data for training and evaluation.  In this paper, we introduce a manually created dataset of multi-turn Socratic advice that is aimed at helping a novice programmer fix buggy solutions to simple computational problems. The dataset is then used for benchmarking the Socratic debugging abilities of a number of language models, ranging from fine-tuning the instruction-based text-to-text transformer Flan-T5 to zero-shot and chain of thought prompting of the much larger GPT-4. The code and datasets are made freely available for research at the link below.\\
\faicon{github}\hspace{0.2cm}\href{https://github.com/taisazero/socratic-debugging-benchmark}{https://github.com/taisazero/socratic-debugging-benchmark}
\end{abstract}

\ccsdesc[500]{Social and professional topics~CS1}
\ccsdesc[500]{Applied computing~Computer-assisted instruction}
\ccsdesc[300]{Computing methodologies~Discourse, dialogue and pragmatics}

\keywords{debugging, introductory programming, Socratic dialogue, language models, benchmark}



\maketitle

\section{Introduction and Motivation}

The increasing demand for computer science (CS) education is straining higher education institutions due to a lack of sufficient instructional staff, often leading to the hiring of undergraduate Teaching Assistants (TAs)~\cite{camp2017generation}. Effective TAs significantly impact student retention rates by providing tailored assistance~\cite{mirza2019undergraduate}, but not all institutions benefit uniformly from their TAs. The challenges of allocating TA time, especially near deadlines or in large classes, are exacerbated by the shortage of K-12 computer science teachers and the lack of appropriate training for K-12 educators interested in teaching CS effectively~\cite{yadav2016expanding}. This further increases TA and peer instruction demand in flipped computer science classrooms~\cite{maher2015flipped}. These factors motivate the automation of teaching tasks by leveraging the capabilities of AI models in understanding and generating language and code.

In Socratic questioning, a teacher assists a learner trying to solve a problem beyond their zone of proximal development~\cite{quintana2018scaffolding}. Language Models (LMs) have been used effectively for generating Socratic questions for math word problems, wherein they leverage the sequential structure of steps that compose the solution~\cite{shridhar-etal-2022-automatic}. Other applications of LMs include automated feedback on student code submissions~\cite{wu2021prototransformer}, or generating programming exercises, unit tests, and code explanations~\cite{macneil_2023_code_explanations}. However, there still remains a substantial gap in leveraging LMs effectively for guiding novice programmers through a coding exercise while maximizing their learning outcomes.

In this paper, we focus on the task of Socratic questioning for debugging~\cite{wilson1987socratic}, or Socratic debugging, defined as a conversation between a knowledgeable programmer and a beginner student who comes for help fixing a buggy solution for a simple computational problem (Section~\ref{sec:task}). To enable the development and evaluation of LM-based instructional agents, we introduce a manually created dataset of dialogues where the main objective is for the student to repair their buggy code themselves by leveraging guidance received from the instructor at every turn (Section~\ref{sec:dataset}). However, as originally observed by~\citet{wilson1987socratic}, "no precise formula, or line of questioning" is needed to achieve the goals of Socratic questioning. Furthermore, depending also on their expectations with respect to the student's abilities, an instructor can often think of multiple ways of guiding the student at any particular turn in the conversation, leading to a very large space of possible dialogues. To facilitate the automatic evaluation and benchmarking of future Socratic questioning systems in terms of their precision and recall, the dataset contributors are asked to provide all alternative utterances that they think could help the student, at every turn in the conversation. This is a complex and cognitively demanding data generation effort, requiring contributors with substantial experience in tutoring beginner programmers. We use the current version of the dataset, containing 151 main conversations, to benchmark the Socratic debugging abilities of two large language models in the GPT family, namely GPT-3.5 and GPT-4 (Section~\ref{sec:experiments}), noticing a large discrepancy in performance in favor of the more recent GPT-4. We conclude the paper with related work and limitations.

\section{Task Definition}
\label{sec:task}

We formulate the Socratic debugging task as a dyadic conversation between a Student and an Instructor. In this scenario, the Student is assumed to be a beginner programmer who has recently started learning how to code in Python. As part of his learning-to-code curriculum, the Student is given a coding problem for which he needs to write a function implementing the specified input-to-output relationship. The Student writes the code for the function, however, the code is buggy and he cannot make progress on his own without help, therefore he seeks help from the Instructor. The Instructor is assumed to be a proficient programmer in Python with experience in teaching novice programmers how to code. When contacted by a Student for help, her main aim is to maximize the learning outcomes by following a Socratic guidance approach through which, over one or more dialogue turns, she helps the students figure out where the bug is and how to fix it on their own.

\subsection{Task Input}
\label{subsec:input}

Since the focus of this work is on generating Socratic guidance and not bug identification or fixing bugs, we assume that the AI agent implementing the Instructor also has access to a description of the bug and one or more bug fixes. The decision to separate Socratic advice generation from bug identification and debugging was motivated by the fact that these subordinate tasks can already be solved efficiently by large LMs with high accuracy. Therefore, at the start of each conversation, we assume the Instructor has access to the {\it problem description}, a number of {\it test cases}, the student's {\it buggy code}, the {\it bug description}, and one or more {\it bug fixes}, as shown below in a sample from our dataset. At each turn in the conversation, the Instructor's task is to generate Socratic guidance in response to the Student's current progress in addressing the bug. Consequently, we assume that the Instructor is also given as input a history of the conversation so far, ending with the last utterance from the student. Shown below is an example ending with the second turn from the student, where the turn number is indicated between brackets.

\begin{newshaded}
\small
\noindent \ding{227} {\bf Problem description}: Write the {\tt factorial(n)} function that computes the factorial {\tt n!} of a natural number {\tt n}, which is defined mathematically as:
\begin{verbatim}
   0! = 1
   n! = n x (n - 1)!
\end{verbatim}
\noindent Additionally, if the input integer {\tt n} is negative the function should return {\tt 0}.
\end{newshaded}
\begin{newshaded}
\noindent\begin{minipage}{.45\columnwidth}
\small
\noindent \ding{227} {\bf Test cases}:
\begin{lstlisting}[basicstyle=\small]
assert factorial(-1) == 0
assert factorial(0) == 1
assert factorial(1) == 1
assert factorial(2) == 2
assert factorial(3) == 6
assert factorial(4) == 24
assert factorial(5) == 120
\end{lstlisting}
\end{minipage}\hfill
\begin{minipage}{.45\columnwidth}
\small
\noindent \ding{227} {\bf Buggy code}:
\begin{lstlisting}[basicstyle=\small]
1. def factorial(n):
2.   if n < 0:
3.     return 0
4.  fact = 1
5.  for i in range(n):
6.    fact = fact * i 
7.  return fact
\end{lstlisting}
\end{minipage}
\end{newshaded}
\begin{newshaded}
\small
\noindent \ding{227} {\bf Bug description}:
On line 6, {\tt fact} is multiplied with {\tt 0} in the first iteration of the for loop. Consequently, at every iteration {\tt fact} stays equal with {\tt 0} instead of being updated to be equal with factorial of {\tt (i + 1)}. Therefore, the function will return {\tt 0}, irrespective of {\tt n}.
\end{newshaded}

\begin{newshaded}
\small
\noindent \ding{227} {\bf Bug fixes}:
\begin{enumerate}
    \item Replace {\tt i} with {\tt (i + 1)} on line 6.
    \item Replace {\tt range(n)} with {\tt range(1,n+1)} on line 5.
\end{enumerate}
\end{newshaded} 
\vspace{0.5em}
At each turn in the conversation, the Instructor's task is to generate Socratic guidance in response to the Student's current progress in addressing the bug. Consequently, we assume that the Instructor is also given as input a history of the conversation so far, ending with the last utterance from the student. Shown below is an example ending with the second turn from the student, where the turn number is indicated between brackets. 
\vspace{0.5em}
\begin{newshaded}
\small
\noindent \ding{227} {\bf Conversation so far}:

\noindent [1] {\sc Student}: Hi! I implemented the factorial but it doesn't work and I do not know why. Can you help?

\noindent[1] {\sc Instructor}: Sure. Can you tell me for what values of {\tt n} it fails and what values it returns in those cases?

\noindent [2] {\sc Student}: For {\tt n = 1} or larger it returns the same value, {\tt 0}.

\noindent [2] {\sc Instructor}: $\langle${\it Socratic guidance}$\rangle$
\end{newshaded}
\vspace{0.5em}
To summarize, the input for the Instructor agent consists of:
\begin{itemize}[leftmargin=*]
    \item[1.] The {\bf problem description}, a number of {\bf test cases}, the student's {\bf buggy code}, the {\bf bug description}, and one or more {\bf bug fixes}.
    \item[2.] The {\bf conversation so far}, ending with the last turn from Student.
\end{itemize}

\subsection{Task Output}

Using the input data described above, the Instructor is expected to generate Socratic guidance appropriate for the current state of the conversation, as shown below.

\begin{newshaded}
\small
\noindent \ding{227} \noindent {\bf Socratic guidance}:

\noindent Main responses:
\begin{itemize}[leftmargin=*]
    \item[] Can you explain, line by line, what happens when running the code for {\tt n} equal to 1?
\end{itemize}
\noindent Alternative responses:
\begin{enumerate}[leftmargin=*]
    \item[1.] However, {\tt fact} is initialized with 1. Where in the code can {\tt fact} change its value to 0?
    \item[2.] What is the first value that is assigned to variable {\tt fact} in line 6?
    \item[3.] What is the first value that is assigned to variable {\tt i} in line 5?
    \item[4.] Can you insert a new line between lines 5 and 6 that prints the values of the variables {\tt fact} and {\tt i}?
    \item[5.] Can you tell me what {\tt range(n)} does?
\end{enumerate}
\end{newshaded}
The example above shows 6 Socratic responses, partitioned into 1 main response and 5 alternative responses. As there may be multiple ways of guiding the student, the Instructor should ideally generate all types of Socratic guidance that are different from each other in non-trivial ways and that cover the entire spectrum of potential information given to the student. For example, the main alternative above provides the minimal amount of information regarding the bug, as it implicitly guides the student to find the bug on their own by tracing the code behavior. This minimal Socratic advice would be suitable for a student who has the ability to find the bug on their own and who only needs to be prodded to mentally run the code and make relevant observations about changes in its runtime state. The 1st alternative provides a bit more focus by asking the student to determine the reason for why {\tt fact} becomes 0 for the first time. The 2nd and 3rd alternatives provide even more Socratic guidance by focusing the student on particular assignments that are determinant for the buggy behavior. The 4th alternative would be suitable for a student who has not yet developed an appropriate understanding of imperative programming as a sequence of changes in program state, and as such he has trouble with tracing the code mentally or on paper. Of all Socratic responses above, the 5th alternative provides the most guidance as it focuses the student on the actual programming construct that is responsible for the bug. While the response with the maximum amount of information would be one providing the bug fix itself, such as "can you replace {\tt range(n)} with {\tt range(1,n+1)} on line 5?", we exclude such responses as they are not in the spirit of the Socratic method.


\section{Benchmark Dataset}
\label{sec:dataset}

To facilitate the development of conversational agents that act under the task definition above, we manually created a dataset of dialogues where a student fixes buggy code on his own by leveraging the Socratic guidance received from an instructor. The dataset is created by specifying the {\it coding problems} \& {\it bugs} (Section~\ref{sec:problems}), followed by the manual contribution of {\it Socratic conversations} \& {\it threads} (Section~\ref{sec:dialogues}) within a browser-based web application (Section~\ref{sec:webapp}).

\subsection{Coding Exercises and Bugs}
\label{sec:problems}

Coding problems are selected to be at a novice level of coding proficiency, such as {\tt Factorial} or {\tt Fibonacci}. Each coding problem is specified through the problem description and the associated test cases. For each problem, one or more buggy implementations are selected, with the constraint that each implementation contains exactly one mistake. The bugs were selected to reflect common types of mistakes that beginner programmers make, such as forgetting that indexing of sequences starts at 0, boundary errors, operator misuse, or misunderstanding of basic programming constructs.
 
A set of 25 programming problems was initially selected by the authors. This was further augmented with 8 coding exercises from the Auckland dataset \cite{ettles_common_2018}, 2 from the Refactory dataset \cite{yang2019refactory}, and 3 from the FalconCode dataset \cite{de_freitas_falconcode_2023}, for a total of 38 problems. The 13 problems collected from external resources were adapted to fit the task definition, such as adding additional unit tests or modifying the exercise to return a solution instead of printing it. The Auckland coding exercises were part of a compulsory assignment in a C programming module for a first-year programming course taught at the University of Auckland in 2016. They were designed to emphasize testing student knowledge of variables, control structures, and arrays. There are over 15 thousand buggy solutions among the collected student submissions. We selected and rewrote in Python 8 problems and 10 corresponding buggy implementations that reflect a diverse set of logical errors. The Refactory dataset contains 5 coding exercises with almost 1,800 buggy Python submissions from 361 students enrolled in a large public university. We selected 2 exercises that require mastery of multiple concepts to solve, such as list operations and search, and 3 corresponding bugs containing logical errors. The FalconCode dataset is a collection of over 1.5 million Python programs from over 2 thousand undergraduate students at the United States Air Force Academy, corresponding to over 800 programming assignments. We selected 3 programming assignments and 4 corresponding bugs to use as a starting point for Socratic conversations. The 3 exercises were selected to be short, e.g. solution in one file, and to not require any file I/O or external dependencies. The corresponding 4 bugs were selected from students who have at least 2 submissions and whose last submission scores a full mark. The bugs were caused by logical errors and were selected to complement types of bugs already included in our dataset.

\begin{table}[t]
    \centering
    \small
    \begin{tabular}{lr|lr}
        \toprule
        Problems & 38 & Dialogues & 151 \\
        Bugs & 57 & Student turns & 1,009\\  
        \hspace{1em} Syntactic & 4 & \hspace{1em} S-utterances & 1,314 \\
        \hspace{1em} Logical & 53 & Instructor turns & 920 \\
        \hspace{2em} Algorithmic & 16 & \hspace{1em} I-utterances & 2,136 \\
        \hspace{2em} Misconception & 32 & All turns &  1,929 \\
        \hspace{2em} Misinterpretation & 9 & \hspace{1em} All utterances & 3,495\\
        \bottomrule
    \end{tabular}
    \caption{Summary of the benchmark dataset: Number of programming problems, bugs, dialogues (including all threads), turns, and total utterances (main and alternatives) for both roles (student and instructor).
    }
    \label{tab:overall_stats}
\end{table}

To get a better sense of the types of bugs included in the dataset, we label each bug with one or more {\it bug categories}. At a high level, there are two major types of bugs, {\it Syntactic} and {\it Logical}. Under the logical bug category, we consider the three broad subcategories introduced by \citet{ettles_common_2018}, {\it Misinterpretation}, {\it Algorithmic}, and {\it Misconception}, listed here in the order they may appear during the problem-solving process. The first type of bugs are caused early on in the process by misinterpretation or misunderstanding of the problem requirements. Algorithmic mistakes are caused by the student using a flawed or incomplete algorithm to solve the problem, such as missing boundary conditions. Finally, misconception bugs reflect a fundamental flaw in programming knowledge, such as forgetting that indexing starts at 0. A breakdown of the 57 bugs across these categories is shown in Table~\ref{tab:overall_stats}. The low number of syntactic bugs is by design and reflects a focus on logical errors, which are much more difficult to fix and hence can benefit more from Socratic guidance. Note that the 4 categories add up to slightly over the total number of bugs because some bugs are labeled with two categories, e.g. a misconception that causes a syntactic error, or a bug that may be caused by either an algorithmic flaw or a misconception. We also associate each bug with a more detailed description where we reference more specific causes or complementary bug type terminology, such as {\it knowledge interference} \cite{spohrer1985bug} or various types of {\it fragile knowledge} \cite{perkins:1986,mccauley_debugging_2008}. The annotated bug types and their detailed descriptions are part of the overall dataset release.

\subsection{Socratic Conversations and Threads}
\label{sec:dialogues}

For each buggy implementation, a main conversation is created, where a fictional Student, the author of the buggy code, interacts with a fictional Instructor. The aim of the instructor is to guide the student to discover the cause of the bug and fix it on his own through Socratic dialogue. The conversation always starts with a student utterance. The instructor and the student then take turns in a dialogue, until the bug is successfully fixed. At each turn, the student may also provide a block of code if he made edits to the code at that turn. Each Main utterance from the main conversation may be followed by one or more Alternative utterances. Given that the aim of this dataset is to benchmark the ability of an artificial Instructor agent to generate Socratic guidance, it is especially important that the Instructor's main and alternative utterances comprehensively explore the entire range of Socratic advice at that point in the conversation. The alternative utterances should be semantically distinct in a non-trivial manner; in particular, they should not be mere paraphrases of the main utterance or of each other. Note that while it is tempting to think of annotating only one Socratic advice at every turn, namely the "optimal Socratic advice", this optimal advice is unknowable due to multiple reasons. The Instructor has insufficient or imprecise relevant knowledge about the Student, ranging from general cognitive abilities, motivation level, programming proficiency, environmental issues, to more transient but still important factors, such as the quality of sleep the night before. The Instructor herself may not be able to optimize a ask as complex as finding the best Socratic advice. These further justify the decision to create all alternative utterances, especially for the Instructor turns. Requiring a comprehensive set of main and alternative instructor utterances is similar to the multiple reference approach introduced by \citet{gupta2019investigating} for improving the evaluation of open-domain dialogue systems. For the Student, alternative utterances may give different or conflicting answers to an Instructor question, reflecting different levels of understanding. Students may give correct or incorrect answers; they may also introduce new bugs when trying to fix the original bug. Once the main conversation ends with the student successfully correcting their code and passing all test cases, the contributors are instructed to further create up to three conversational threads.

Upon inspection of the conversations created manually, we discovered that one contributor had used a vending machine as an analogy to guide the user to conclude that {\tt print} was not the same as {\tt return}. While analogies can substantially enhance the impact of Socratic questioning, it can lead to an open-ended range of alternatives, as the number of possible analogies is virtually infinite. Since our aim is to create a dataset that can be used to estimate both the recall and precision of a Socratic guidance generator, at this stage we decided to require that Socratic utterances be {\it literal}, leaving the generation of figurative utterances as a direction for future work.

\begin{table*}[!h]
    \centering
    \begin{tabular}{crrr|rrr|rrr|rrr}
        \toprule
         \multicolumn{1}{c}{\textbf{Language}} & \multicolumn{3}{c}{\textbf{Manual}} & \multicolumn{3}{c}{\textbf{BLEU-4}} & \multicolumn{3}{c}{\textbf{BERT F1}} & \multicolumn{3}{c}{\textbf{Rouge-L}} \\
         \multicolumn{1}{c}{\textbf{Model}} & P & R & F1 & P & R & F1 & P & R & F1 & P & R & F1 \\
        \midrule
        GPT-3.5 & 22.8 & 21.7 & 22.2 & 3.2 & 1.7 & 2.0 & 56.0 & 43.5 & 48.9 & 21.0 & 14.3 & 17.0 \\
        \hspace{0.4cm}  + CoT & 18.6 & 5.5 & 8.5 & 2.3 & 0.8 & 1.1 & 61.7 & 35.8 & 41.6 & 20.3 & 9.7 & 12.0 \\
        GPT-4 & 42.5 & 42.7 & 42.6 & 3.2 & 4.3 & 3.6 & 35.4 & 62.6 & 45.2 & 14.1 & 23.3 & 17.6 \\
        \hspace{0.4cm}  + CoT & 38.2 & 57.5 & 45.9 & 0.9 & 4.8 & 1.4 & 12.6 & 64.8 & 19.5 & 5.2 & 26.6 & 8.1 \\
        \bottomrule
    \end{tabular}
    \caption{Baseline evaluation of GPT-3.5 (\texttt{gpt-3.5-turbo}) and GPT-4 on our benchmark dataset. The "+ CoT" row represents the evaluation of the language model using the Chain of Thought (CoT) prompting approach. We report the Precision (P), Recall (R), and F1 for the manual evaluation, and BLEU-4, BERT F1, and Rouge-L for the automatic evaluation. All results are percentages (\%).}
    \label{tab:evaluation}
\end{table*}
Overall, the dataset contains a total of 151 dialogues and 3,495 utterances, all created by ten contributors with extensive experience in CS education as instructors, teaching assistants, or tutors. The right half of Table~\ref{tab:overall_stats} shows more detailed statistics in terms of the total numbers of student/instructor turns and utterances. We compute the human inter-annotator agreement (ITA) on a subset of 5 dialogues composed of 24 turns containing over 73 instructor utterances. To compute ITA, we ask one data contributor $C$ to write instructor utterances (main and alternative) given the conversation so far as input, for each of the 5 evaluation dialogues. These dialogues have already been completed by other data contributors $\neg{C}$. We then perform manual evaluation using the procedure detailed in Section~\ref{subsec:manual_eval}, where we assess the turns contributed by the data contributor $C$ against the ground-truth utterances contributed by the other data contributors $\neg{C}$. The ITA evaluation results in $P = 77.4$, $R = 46.1$, and $F_1 = 57.8$. This is substantially higher than the best-performing system shown in Table~\ref{tab:evaluation}, especially in terms of precision, demonstrating that a human data contributor can reliably write high-quality Socratic utterances. Furthermore, $82.1\%$ of the correct Socratic utterances written by $C$ were semantically equivalent to an utterance contributed by $\neg{C}$ indicating strong coverage of the benchmark dataset. We observe a lower recall indicating that a data contributor on their own may not be as comprehensive as all contributors combined in capturing the wide array of Socratic utterances listed in the benchmark.

\subsection{Web Application}
\label{sec:webapp}

To streamline and standardize the collection of Socratic dialogues and code edits for each input problem description and buggy implementation, we developed a 7-page web application using the Streamlit and gsheetsdb libraries. The application guides contributors through selecting a bug, creating initial and conversational threads, and reviewing and submitting their work. During the process, contributors can add main and alternative utterances, undo actions, and edit the chat history. The application also allows importing and exporting dialogues in a standardized form for review.

\section{Eliciting Socratic Advice from Large Language Models}
\label{sec:lms}

We evaluate the GPT-3.5~\cite{chatgpt} and GPT-4~\cite{gpt4} language models in terms of their capacity to generate, at each instructor turn, Socratic utterances that match those contributed in the benchmark dataset. Each test example is composed of an input prompt to the language model containing: a steering prompt for Socratic questioning adapted from the GPT-4 blog post~\cite{oai_blog}, the problem description, the buggy code, the bug description, the bug fixes, the unit tests, the dialogue history so far, and an instruction to the language model to generate all possible semantically distinct Socratic utterances as the instructor.


The list of utterances generated by the LM is then used to estimate precision and recall. After conducting a preliminary, qualitative evaluation of various prompts and instructions we select the prompts and instructions used in this paper.



For the GPT models, we use the following instruction in the standard zero-shot setting experiment where the language model is given an instruction without any examples:
\begin{newshaded}
\small
     Respond to the user with all possible distinct Socratic utterances that guide the user to discover and fix the bug described between `<bug\_desc>' and `</bug\_desc>'. Student code is written between `<code>' and `</code>' throughout the conversation. Utterances that have the same meaning but different words are considered duplicates. Assume that the student has run the test cases.
\end{newshaded}

Chain of Thought (CoT)~\cite{wei2022chain} is a language model prompting method that decomposes the problem into intermediate steps that lead to a final answer. We utilize the CoT prompting approach to decompose the Socratic utterance generation problem into two steps. The first step involves reasoning about the learner's misconceptions and other reasons that may have caused the learner to write the buggy code initially and continue to impede the learner from fixing the bug. In this step, the language model lists reasons and possible misconceptions given the programming problem, buggy code, bug description, bug fixes, and the conversation so far and is instructed to do so given the following prompt:
\begin{newshaded}
\small
     Given the dialogue so far, what are all the possible reasons or misconceptions if any that the user still has that impede them from fixing the bug? Do NOT list Socratic questions. If the bug is already fixed, say ``There are no remaining misconceptions or reasons that impede the user from fixing the bug, as they have already identified and corrected their code."'
\end{newshaded}
\vspace{0.5em}
In the second step, the LM is asked to utilize the dialogue so far and the list of possible reasons and misconceptions from the previous step to generate a list of Socratic utterances.
We also experiment with fine-tuning open-source FLAN-T5 models~\cite{chung2022scaling} (small, base, and large) using a learning rate of $1e-5$, an Adam optimizer~\cite{kingma2014adam} and an effective batch size of 32 for 20 epochs. However, the fine-tuned LMs performed very poorly on a held-out test set of 16 dialogues around unseen programming exercises. This low performance is likely due to a low exposure of the FLAN-T5 models to code dialogues and Socratic questioning prior to fine-tuning.

\section{Experimental Evaluations}
\label{sec:experiments}

In all experiments, LM outputs are generated using a greedy decoding setting (i.e. temperature = 0). We set a maximum generated token threshold of 1,024 and do not apply any frequency or presence penalties. We perform manual evaluation of the LM generations for a subset of problems, and automatic evaluations for all problems in the benchmark dataset.

\subsection{Manual Evaluation}
\label{subsec:manual_eval}

We aim to estimate the performance of GPT-3.5 and GPT-4 by manually assessing the quality of their generated instructor utterances. At each instructor dialogue turn, we manually examine each LM utterance to determine if it is an appropriate Socratic utterance at that turn. We sample a total of 149 instructor utterances composing 43 instructor turns across 11 dialogues from the benchmark. Using the example listed in Section~\ref{subsec:input}, during the second instructor turn a good-matching generated utterance example is: ``How does the range function work in your loop, and what values does it generate for {\tt i}?" because it is semantically close with the ground truth utterance: ``Can you tell me what {\tt range(n)} does?".  If the LM utterance is good but not present in our dataset, we mark it as missing to compute an overall upper bound on recall for the dataset itself. These missing alternatives can later be used to augment the dataset. If the LM output is not good, it is considered a false positive (FP), which decreases the precision of the LM. For each alternative in the benchmark dataset at that turn, we check if it is missing from the list of LM utterances. If missing, it is considered a false negative (FN), which decreases the recall of the LM. If the dataset utterance semantically matches any of the LM utterances, it is considered a true positive (TP). If the LM generates two or more paraphrases of the same Socratic guidance, they are considered as one Socratic utterance. The precision (P), recall (R), and their harmonic mean (F1) presented in Table~\ref{tab:evaluation} highlight GPT-4's superior performance over GPT-3.5 in generating relevant and diverse Socratic utterances. We emphasize GPT-3.5's poor precision as it tends to generate many poor Socratic questions (93 FP) compared to GPT-4 (41 FP) that may contain keywords in common with a ground truth utterance but are irrelevant. We also observe lower GPT-3.5 performance when using the CoT approach. This is due to its poor ability in listing possible reasons and misconceptions. On the other hand, we observe an increase in recall at the expense of precision when using CoT when using GPT-4. This is due to GPT-4 generating more Socratic utterances focused on addressing various possible reasons or misconceptions typically while generating utterances that have already been asked or answered, or too early where the student is not yet aware of the issue. Furthermore, the CoT approach yielded a 10\% increase in good utterances that are not present in the dataset. In addition, we compute the recall for our benchmark dataset during human evaluations, obtaining a value of $75.5$. This score suggests that most of the high-quality Socratic utterances generated by LMs are effectively captured within the dataset by contributors, further validating the dataset's usefulness for benchmarking purposes. Human evaluation was conducted by 3  contributors with extensive experience in CS education. We compute manual evaluation agreement (MEA) across the 3 evaluators using a sample of 10 dialogues containing over 35 turns and 77 utterances obtaining a value of $84.6\%$ agreement.


\subsection{Qualitative Analysis}
\label{subsec:qualitative_eval}
GPT-3.5 often refers to example cases in problem descriptions when asking for expected output, while GPT-4 does this less frequently. GPT-4, however, generates more diverse and comprehensive Socratic utterances, producing 32 good utterances not found in the dataset, three times more than GPT-3.5's 10. Also, GPT-3.5 generates a higher number of poor utterances, with over 96 compared to GPT-4's 42. When the CoT approach is used, GPT-4 generates more comprehensive Socratic utterances, but at the cost of overall quality, as it produces more duplicates and poor utterances. Below are cherry and lemon-picked LM output samples given a buggy code containing a return statement within the for loop leading to early termination. Good responses are bolded and colored in \tcg{\textbf{green}} while poor utterances are italicized and colored in \tcr{\textit{red}}.

GPT-4's first 3 utterances demonstrate good Socratic utterances offering distinct levels of guidance. The first directs the student's attention to the erroneous code line, the second probes the student's comprehension of the 'return' keyword, and the third provides the most guidance by questioning the influence of indentation on code execution. GPT-3.5's second utterance is an example of poor guidance. Ineffective guidance falls into four categories: irrelevant, repeated, overly direct, and premature utterances. \textbf{\textit{Irrelevant utterances}}, such as GPT-3.5's third, shift focus from the actual bug, leading the student away from the loop and its termination and can lead to confusion. GPT-3.5 produced 53 such irrelevant utterances, significantly more than GPT-4's 8. \textbf{\textit{Repeated}} utterances, like GPT-3.5's fourth one, are already asked or answered. Utterances that are too \textbf{\textit{direct}}, such as GPT-4's fourth one, disclose the bug too early, reducing the learning challenge. Finally, \textbf{\textit{premature}} utterances guide learners to code changes before they identify the issue, potentially causing confusion, as seen in GPT-3.5's second utterance.
\vspace{0.5em}
\begin{newshaded}
\small
\noindent \ding{227} \noindent {\bf Conversation So Far}:
\newline
\noindent [1] {\sc Student}: Help, I'm stuck.
\newline
\noindent [1] {\sc Instructor}: What should your code be doing?
\newline
\noindent [2] {\sc Student}: It's only returning the first number in the list, not all.
\newline
\noindent [2] {\sc Instructor}: Okay, let's look at your loop. Do you see anything that could be causing it to end early?
\newline
\noindent [3] {\sc Student}: No. I'm iterating through the whole list of numbers.
\newline
\noindent \ding{227} \noindent {\bf Sample GPT-4 Outputs}:
\begin{enumerate}
  \setlength{\itemsep}{2pt}
  \setlength{\parskip}{0pt}
 \item \tcg{\textbf{Can you identify which part of your code is responsible for returning the final result?}}
 \item \tcg{\textbf{What happens when the `return' statement is executed inside the loop?}}
 \item \tcg{\textbf{How does the placement of the `return' statement affect the execution of your code?}}
\item  \tcr{\textit{How can you modify the position of the `return' statement to ensure that the entire list is processed before returning the result?}} \tcb{\textit{(direct)}} 
\end{enumerate}

\noindent \ding{227} \noindent {\bf Sample GPT-3.5 Outputs}:
\begin{enumerate}
  \setlength{\itemsep}{2pt}
  \setlength{\parskip}{0pt}
    \item \tcg{\textbf{What does the `return' statement do?}}
    \item \tcr{\textit{How can you modify the code to return all odd numbers in the list?}} \tcb{\textit{(premature)}} 
    \item \tcr{\textit{What happens when the if statement is true?}} \tcb{\textit{(irrelevant)}} 
    \item \tcr{\textit{Can you identify any issues with the loop?}} \tcb{\textit{(repeated)}} 
\end{enumerate}
\end{newshaded}
\vspace{0.5em}

\subsection{Automatic Evaluation}

Following prior work in Socratic sub-question generation~\cite{shridhar-etal-2022-automatic}, we compute the similarity between an LM utterance and a ground truth utterance in the dataset using BLEU~\cite{papineni-etal-2002-bleu} for n-gram overlap, BERT F1 Score~\cite{bertscore} for semantic similarity based on the DeBERTa language model~\cite{he2020deberta}, and Rouge-L~\cite{lin-2004-rouge} for n-gram overlap based on Longest Common Subsequence (LCS) between generated and reference instructor utterances. Rouge-L is included for its flexibility in evaluating text similarity and capturing overall structure and content better than BLEU-4. BERTScore is included to handle paraphrases. Given a set of $m$ LM-generated utterances and $n$ manually created utterances, we create a complete bipartite graph between the two sets, with a total of $mn$ edges, where the weight of each edge is computed using one of the text similarity measures above. We then apply Edmond's Blossom algorithm \cite{galil1986efficient} for finding the maximum matching in this bipartite graph. This ensures that each manual utterance is matched with at most one LM utterance, effectively prohibiting semantically equivalent LM utterances from artificially increasing the evaluation measures. The number of true positives $TP$ is computed by summing up the weights of all edges found in the optimal matching. Given that the weights are similarity scores in $[0, 1]$, if an LM utterance $u$ is matched with a manual utterance $v$ for a similarity weight of $s(u, v)$, the remaining weight mass of $1 - s(u, v)$ is considered to contribute towards the total number of false positives $FP$. Any unmatched LM utterance is considered to contribute the maximum of 1 towards the $FP$ total. Overall, it can be shown that this results in $FP = m - TP$. The number of false negatives is computed in an analogous way, resulting in $FN = n - TP$. Consequently, precision is $P = TP / m$ and recall is $R = TP / n$.

Table~\ref{tab:evaluation} displays the results of the automatic evaluation for GPT-3.5 and GPT-4 on the full benchmark dataset. Automatic metrics do not correlate with manual evaluation in GPT-4's CoT experiment, likely due to CoT introducing new utterances absent from the dataset more often and low lexical and semantic overlap between actual and generated utterances. For example ``What should your code be doing" semantically matches ``What do you expect your for loop to do?" but it has no lexical overlap and scores less than 0.4 on BERT F1. Low automatic scoring may happen irrespective of the utterance's relevance or usefulness, underlining the importance of manual evaluation for this task.



\section{Related Work}
Scaffolding is a process in which a learner achieves a goal through guided efforts~\cite{wood1976role}. Socratic questioning (SQ), a part of scaffolding theory~\cite{wood1976role}, assists learners in solving problems beyond their zone of proximal development~\cite{vygotsky2012thought} by prompting them with questions that guide them towards the solution. Recent work has shown that SQ is effective in enhancing learning gains in code comprehension tasks~\cite{tamang2021_socratic}. Various techniques, including LMs, were proposed for generating automated hints to assist novices in programming~\cite{lazar2017automatic,rivers2017data,lee2018vida,mcbroom2021survey,juho_2023_error_messages}. However, fixing misconceptions is often time-consuming~\cite{perkins1986fragile}. Chen et al.~\cite{chen-etal-2011-exploring} curate a computer science tutoring corpora and classify tutor utterances into procedural, declarative, prompts, and feedback utterances. There is limited work on corpora focused on guiding students to discover and fix bugs using SQ. Recent studies employ Pairwise comparison tests on BlenderBot and GPT-3 to assess their educational abilities, showing these models fall short of human teachers in understanding, aiding, and mimicking teacher-like dialogue in tutoring contexts~\cite{brown2020language,roller-etal-2021-recipes, gpt3_tutor_2022}.
\section{Conclusions}
This paper presents a dataset of expert-curated Socratic conversations where instructors assist novice debuggers in fixing buggy programs. The dataset serves as a benchmark for evaluating the Socratic debugging capabilities of LMs. Although GPT-4 outperforms GPT-3.5, its precision and recall remain below human expert levels, highlighting the need for further research.

\section{Acknowledgements}
We would like to thank Laurel Powel, Khyati Mahajan, and Mohsen Dorodchi for their contributions to a previous version of the dataset \cite{al-hossami-etal-2023-socratic}. This research was partly supported by the United States Air Force (USAF) under Contract No. FA8750-21-C-0075. 
\bibliographystyle{ACM-Reference-Format}
\bibliography{sigcse24}


\end{document}